\documentclass[runningheads]{llncs}

\usepackage{graphicx}
\usepackage{hyperref}
\usepackage{color}

\begin{document}

\title{Intergrated Segmentation and Detection Models for Dentex Challenge 2023}

\author{Lanshan He\inst{1}\orcidID{0009-0009-6093-8803} \and
    Yusheng Liu\inst{1}\orcidID{0009-0004-2624-9223} \and
    Lisheng Wang\inst{1}\orcidID{0000-0003-3234-7511}}

\authorrunning{L. He et al.}

\institute{Institute of Image Processing and Pattern Recognition, Department of Automation,
    Shanghai Jiao Tong University, Shanghai 200240, People's Republic of China \\
    \email{\{xyzlancehe,lswang\}@sjtu.edu.cn}}

\maketitle

\begin{abstract}
    Dental panoramic x-rays are commonly used in dental diagnosing. With the development of deep learning,
    auto detection of diseases from dental panoramic x-rays can help dentists to diagnose diseases more efficiently.
    \href{https://dentex.grand-challenge.org/}{The Dentex Challenge 2023} is a competition
    for automatic detection of abnormal teeth along with their enumeration ids from dental panoramic x-rays.
    In this paper, we propose a method integrating segmentation and detection models to detect abnormal teeth as well as
    obtain their enumeration ids.
    Our codes are available at \url{https://github.com/xyzlancehe/DentexSegAndDet}.

    \keywords{Dental Panoramic X-ray  \and Tooth Segmentation \and Tooth Detection  \and Disease Detection.}
\end{abstract}

\section{Introduction}
Dental panoramic x-ray is a commonly used imaging technique in dental diagnosing.
Howerver, manual diagnosis on dental panoramic x-ray is time-consuming and requires professional knowledge.
There have been many deep learning based methods for automatic segmentation or detection from dental panoramic x-rays.
With the help of these methods, the efficiency of dental diagnosing on panoramic x-rays can be greatly improved.
To make further progress, \href{https://dentex.grand-challenge.org/}{The Dentex Challenge 2023}\cite{hamamci2023dentex} is presented,
which requires abnormal teeth to be detected, and their corresponding enumeration ids in FDI notation to be obtained simultaneously.
The challenge has released a dataset containing hierarchically labeled dental panoramic x-rays.
Based on the challenge and the dataset, we propose a method that integrates several segmentation and detection models to detect abnormal teeth,
and obtain their disease labels and enumeration ids.

\section{Method}
There are many powerful models for segmentation and detection tasks on 2D images, such as
U-Net\cite{ronneberger2015unet}, a classic segmentation model widely used in medical images,
and DETR\cite{carion2020detr}, an end-to-end object detection model based on transformer architecture.
These models usually predicts one class label for each pixel or bounding box.
By modifying the structure of detection head, we can obtain multi-class labels from a detection model,
such as HierarchicalDet\cite{hamamci2023diffusion}.
\par
In this paper, we break down the task into two subtasks: tooth detection and disease detection, and then
merge their results to obtain the final output.
Each subtask only needs to predict a single class label for each pixel or bounding box.
Thus, we can utilize those powerful models and do not have to modify their structures greatly.
\par
\begin{figure}
    \includegraphics[width=\textwidth]{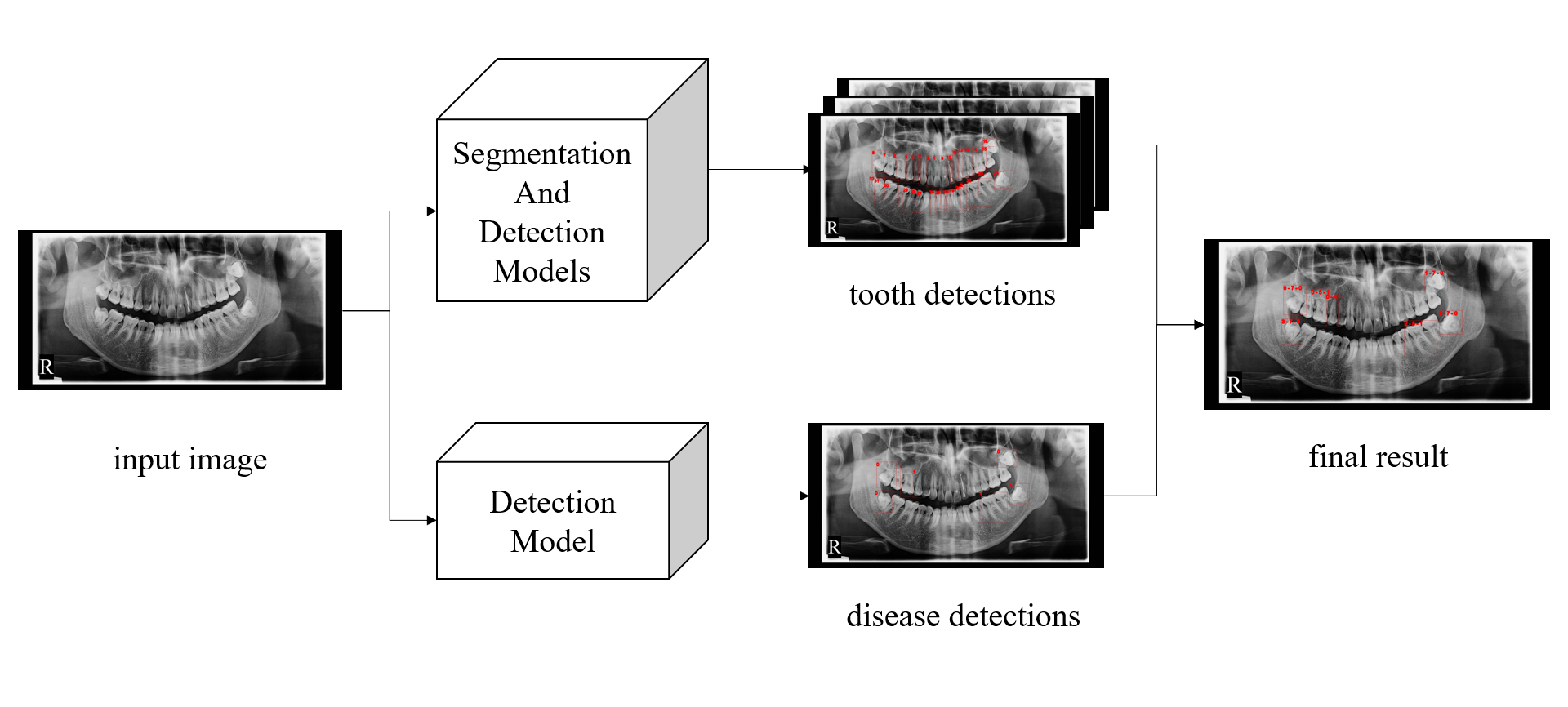}
    \caption{An overview of our pipeline} \label{fig1}
\end{figure}
As shown in Fig.~\ref{fig1}, the pipeline of our method mainly consists of three steps: tooth detection, disease detection and label matching.

\paragraph{Tooth Detection}
We use both segmentation models and detection models to perform tooth detection.
Segmentation models produce pixel-wise tooth segmentation masks, which are then converted to bounding boxes.
Detection models directly produce tooth bounding boxes.
The result of each model forms a dictionary, containing the detected tooth enumeration ids and their corresponding bounding boxes.

\paragraph{Disease Detection}
We use a detection model to perform disease detection on the image, getting diseased teeth's labels and their bounding boxes.

\paragraph{Label Matching}
For each diseased tooth, we try to match its enumeration id from the tooth detection dictionaries. Each enumeration id is voted
by the IOU of its tooth bounding box and the diseased tooth's bounding box. The enumeration id with the highest vote is chosen as the
final enumeration id of the diseased tooth.

\subsection{Preprocessing}
The dataset contains three labeled subsets. The first subset contains quadrant labels, where each quadrant object is labeled with quadrant id.
The second subset contains tooth labels, where each tooth object is labeled with quadrant id and enumeration id in the quadrant.
The third subset contains disease labels, where each diseased tooth object is labeled with quadrant id, enumeration id and disease id.
The labels of each subset are in a json file, slightly modified from COCO dataset's annotation format.
\par
For training the detection models, we convert the format of the dataset, arranging them into both COCO format and YOLO format.
Meanwhile, we convert the second subset's labels from quadrant id and enumeration id in quadrant into global enumeration id, ranging from 1 to 32.
And for the third subset, we only keep the disease id, ignoring the quadrant id and enumeration id.
\par
For training the segmentation models, we draw segmentation masks for each x-ray image from the json file according to the "segmentation" field of each object.
\par
The rest of the preprocessing conforms the common practice of the downstream models. For the segmentation models, we resize the images and masks to 256x256, and apply
random vertical flip, random crop and random rotation. For the detection models, use the default settings of the downstream models.

\subsection{Tooth Detection}
Both detection models and segmentation models are used to detect teeth. Furthermore, we use two different strategies
to perform tooth segmentation: segmentation on the whole image and segmentation on each quadrant.
\subsubsection{Detection Model}
The detection model we use is DINO\cite{zhang2022dino,li2022dn,liu2022dabdetr}, a detection model based on DETR with faster convergence speed.
We choose resnet50 as the backbone here, and pretrained weights on COCO dataset are used.
\subsubsection{Segmentation Model}
The segmentation models we use are U-Net\cite{ronneberger2015unet} and U-Net with SE-blocks\cite{senet}(SE U-Net).
\par
For segmentation on the whole image, we train models on each whole image of the second subset, where each model directly predicts
32 classes, corresponding to the 32 enumeration ids.
\par
For segmentation on each quadrant, we first use DiffusionDet\cite{chen2022diffusiondet} to detect quadrants, which is trained on the first subset.
Then, we run the trained DiffusionDet model on the second subset,
cropping each quadrant from the whole images and masks according to the detected quadrant bounding boxes.
During this, we convert the masks' value of teeth in current quadrant to 1~8.
Additionally, as the cropped quadrant area may contain teeth from other quadrants, we assign masks' value of all the teeth that
do not belong to current quadrant to 9.
Finally, we train models on each cropped quadrant, where each model predicts 9 classes,
corresponding to the 8 enumeration ids in the quadrant and the teeth not in the current quadrant.
\par
To obtain bounding boxes from the segmentation masks, we first filter the connected components of the masks, where
only at most one largest connected area is kept for each label value, ignoring any other smaller connected areas.
Then, we use the bounding boxes of the connected components as the bounding boxes of the corresponding tooth.
At inference time, each quadrant-segmentation model shares the same trained DiffusionDet model to crop quadrants from the whole image.
Then, the segmentation model is run on each cropped quadrant. After that, bounding boxes are calculated in the quadrant areas and are restored to the whole image according to the quadrants' bounding boxes.
\par
Each detection model and segmentation model works separately, and their results are stored in dictionaries, ready for voting in the label matching step.

\subsection{Disease Detection}
The disease detection model we use is DINO and YOLOv8.
Pretrained weights on COCO dataset are used.
\par
For DINO, we choose swin-transformer\cite{liu2021swin} as the backbone here.
Before the DINO model is trained on the third subset, it is fisrt trained on the second dataset to learn tooth detection.
After that, the weights are kept except the head layer, and training is continued on the third subset, which we think may help to improve performance.
For YOLOv8, we train the model on the third subset directly.
\par
At inference time, we ensemble the results of DINO and YOLOv8 using weighted boxes fusion\cite{wbf}, which takes place before the label matching step.

\subsection{Label Matching}
After tooth detection and disease detection, we merge the results of the two steps to obtain the final output.
For each diseased tooth, we caculate the IOU of its bounding box and each tooth's bounding box in the tooth detection dictionaries.
Each IOU is accumulated to the corresponding enumeration id, and the enumeration id with the highest vote is chosen as the final enumeration id of the diseased tooth.
We also assign different weights to different tooth detection results. Specifically, the results of detection models are assigned higher weights
because they predict bounding boxes directly.

\subsection{Postprocessing}
We apply some postprocessing after the label matching step. First, we convert the enumeration ids to FDI notation, which is the required output format.
Then, we filter out results with duplicate labels. If there are multiple results with the same tooth and disease, we keep the label with the highest confidence score.
Finally, we filter the results by some prior knowledge. For example, the impacted teeth are always the 8th tooth in each quadrant.
As a result, we can filter out some results that do not conform to the prior knowledge.

\section{Result Analysis}
\subsection{Quantitative Results}

\begin{figure}
    \includegraphics[width=\textwidth]{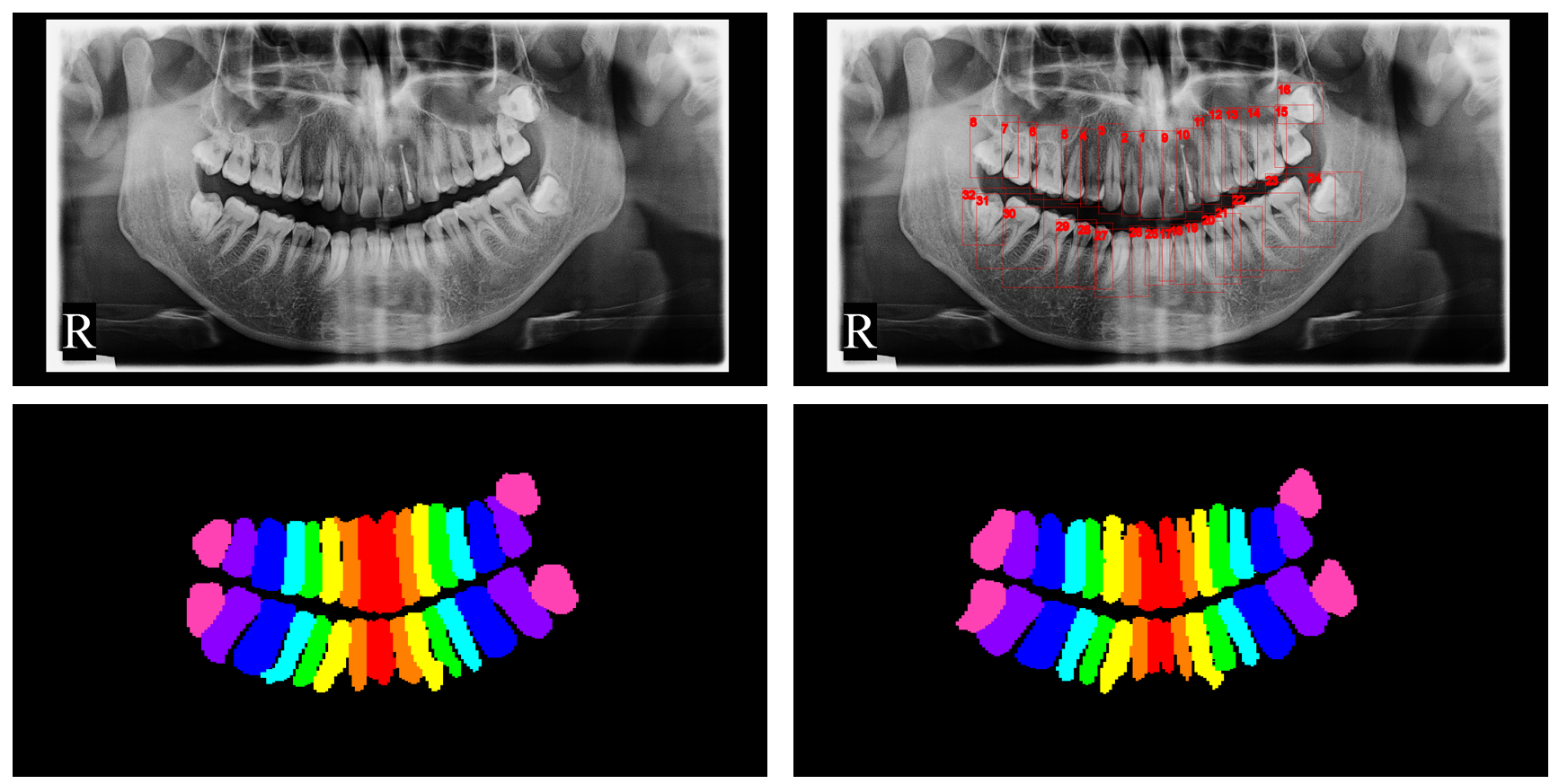}
    \caption{Origin image, detection and segmentation result on whole image} \label{fig2}
\end{figure}

\begin{figure}
    \includegraphics[width=\textwidth]{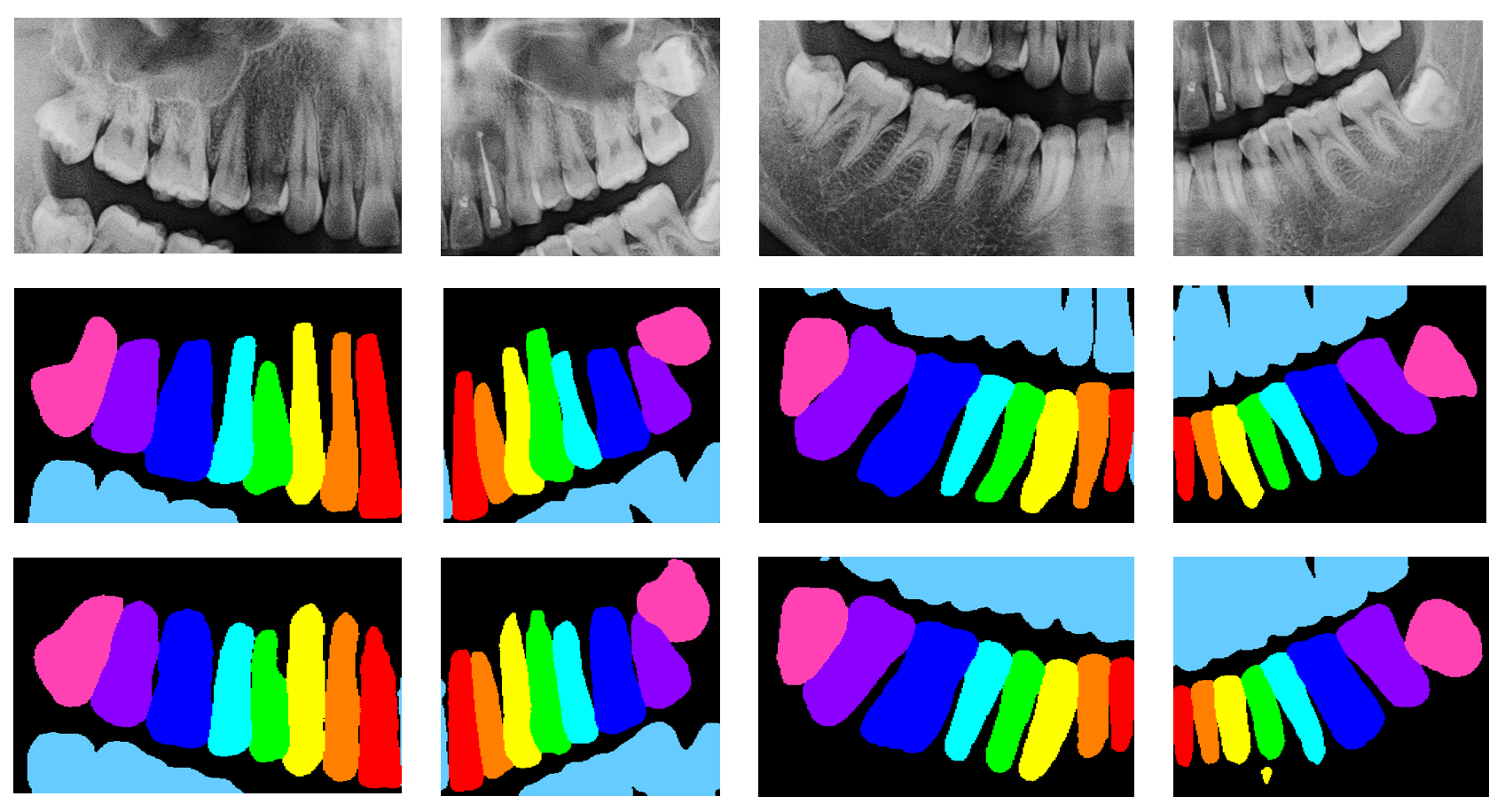}
    \caption{Cropped quadrants and their segmentation results} \label{fig3}
\end{figure}

\begin{figure}
    \includegraphics[width=\textwidth]{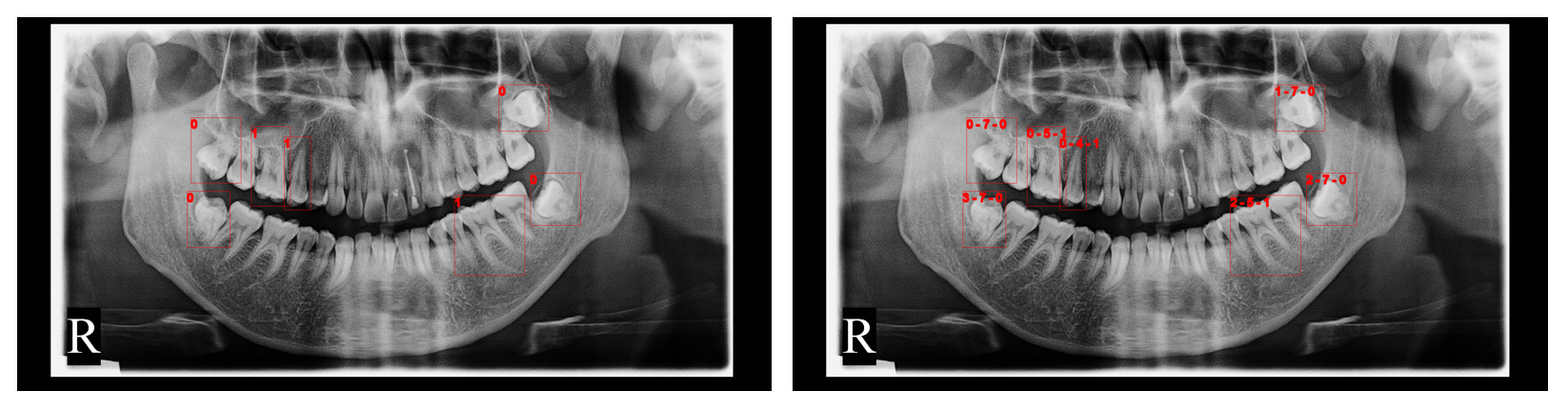}
    \caption{Disease detetcions and final labels} \label{fig4}
\end{figure}

Quantitative results are shown in Fig. \ref{fig2} \ref{fig3} \ref{fig4}.
\par
In Fig. \ref{fig2}, we show the origin image, teeth bounding boxes from the teeth detection model, and teeth segmentation masks from the teeth segmentation models, respectively.
The bottom-left segmentation mask is from U-Net, and the bottom-right segmentation mask is from SE U-Net. We use eight colors to draw the masks while they are actually 32 classes.
\par
In Fig. \ref{fig3}, we show the cropped quadrants by the quadrant detection model in the first line, and their segmentation results in the following two lines.
The second line is from U-Net, and the third line is from SE U-Net.
Though there may be noises in the segmentation results, most of them can be removed by filtering the connected components.
Meanwhile, the segmentation results do not have to be quite accurate, as they are converted to bounding boxes later.
\par
In Fig. \ref{fig4}, we show the disease detection results of the image, and the final labels after label matching, where the enumeration ids are converted to FDI notation(indexes begin from 0).
Here we only draw disease instances whose scores are higher than 0.3. There will be much more instances if lowering the threshold,
which may be helpful for getting higher metrics.
As shown in these figures, our method can detect diseased teeth, and match their enumeration ids correctly.

\subsection{Qualitative Results}
The performance of DINO, YOLOv8 and the fused results of the disease detection models are shown in Table~\ref{tab1}.
\begin{table}
    \caption{Metrics of disease detection.}\label{tab1}
    \centering
    \begin{tabular}{|l|c|c|c|c|}
        \hline
        Model  & $AR$              & $AP$              & $AP_{50}$         & $AP_{75}$         \\
        \hline
        DINO   & 0.544             & 0.343             & 0.508             & 0.411             \\
        YOLOv8 & 0.530             & 0.352             & 0.539             & 0.399             \\
        Fused  & {\bfseries 0.592} & {\bfseries 0.371} & {\bfseries 0.543} & {\bfseries 0.447} \\
        \hline
    \end{tabular}
\end{table}
It is shown that the fused results of the two models achieve better performance than each single model.
\par

Our method's evaluation metrics on the challenge's final test set are shown in Table~\ref{tab2}.
\begin{table}
    \caption{Evaluation metrics on the test set.}\label{tab2}
    \centering
    \begin{tabular}{|l|c|c|c|c|}
        \hline
        Label type  & $AR$  & $AP$  & $AP_{50}$ & $AP_{75}$ \\
        \hline
        Quadrant    & 0.753 & 0.474 & 0.679     & 0.585     \\
        Enumeration & 0.681 & 0.353 & 0.511     & 0.421     \\
        Disease     & 0.592 & 0.371 & 0.543     & 0.447     \\
        \hline
    \end{tabular}
\end{table}

\bibliographystyle{splncs04}
\bibliography{main}

\begin{thebibliography}{10}
\providecommand{\url}[1]{\texttt{#1}}
\providecommand{\urlprefix}{URL }
\providecommand{\doi}[1]{https://doi.org/#1}

\bibitem{carion2020detr}
Carion, N., Massa, F., Synnaeve, G., Usunier, N., Kirillov, A., Zagoruyko, S.:
  End-to-end object detection with transformers. In: European conference on
  computer vision. pp. 213--229. Springer (2020)

\bibitem{chen2022diffusiondet}
Chen, S., Sun, P., Song, Y., Luo, P.: Diffusiondet: Diffusion model for object
  detection (2022)

\bibitem{hamamci2023diffusion}
Hamamci, I.E., Er, S., Simsar, E., Sekuboyina, A., Gundogar, M., Stadlinger,
  B., Mehl, A., Menze, B.: Diffusion-based hierarchical multi-label object
  detection to analyze panoramic dental x-rays. arXiv preprint arXiv:2303.06500
   (2023)

\bibitem{hamamci2023dentex}
Hamamci, I.E., Er, S., Simsar, E., Yuksel, A.E., Gultekin, S., Ozdemir, S.D.,
  Yang, K., Li, H.B., Pati, S., Stadlinger, B., et~al.: Dentex: An abnormal
  tooth detection with dental enumeration and diagnosis benchmark for panoramic
  x-rays. arXiv preprint arXiv:2305.19112  (2023)

\bibitem{senet}
Hu, J., Shen, L., Sun, G.: Squeeze-and-excitation networks. CoRR
  \textbf{abs/1709.01507} (2017), \url{http://arxiv.org/abs/1709.01507}

\bibitem{li2022dn}
Li, F., Zhang, H., Liu, S., Guo, J., Ni, L.M., Zhang, L.: Dn-detr: Accelerate
  detr training by introducing query denoising. In: Proceedings of the IEEE/CVF
  Conference on Computer Vision and Pattern Recognition. pp. 13619--13627
  (2022)

\bibitem{liu2022dabdetr}
Liu, S., Li, F., Zhang, H., Yang, X., Qi, X., Su, H., Zhu, J., Zhang, L.:
  {DAB}-{DETR}: Dynamic anchor boxes are better queries for {DETR}. In:
  International Conference on Learning Representations (2022),
  \url{https://openreview.net/forum?id=oMI9PjOb9Jl}

\bibitem{liu2021swin}
Liu, Z., Lin, Y., Cao, Y., Hu, H., Wei, Y., Zhang, Z., Lin, S., Guo, B.: Swin
  transformer: Hierarchical vision transformer using shifted windows (2021)

\bibitem{ronneberger2015unet}
Ronneberger, O., Fischer, P., Brox, T.: U-net: Convolutional networks for
  biomedical image segmentation. In: Medical Image Computing and
  Computer-Assisted Intervention--MICCAI 2015: 18th International Conference,
  Munich, Germany, October 5-9, 2015, Proceedings, Part III 18. pp. 234--241.
  Springer (2015)

\bibitem{wbf}
Solovyev, R., Wang, W., Gabruseva, T.: Weighted boxes fusion: Ensembling boxes
  from different object detection models. Image and Vision Computing pp.~1--6
  (2021)

\bibitem{zhang2022dino}
Zhang, H., Li, F., Liu, S., Zhang, L., Su, H., Zhu, J., Ni, L.M., Shum, H.Y.:
  Dino: Detr with improved denoising anchor boxes for end-to-end object
  detection (2022)

\end{thebibliography}

\end{document}